\documentclass[10pt,twocolumn,letterpaper]{article}

\usepackage{cvpr}              

\usepackage{graphicx}
\usepackage{xspace}
\usepackage{booktabs}
\usepackage{multirow}
\usepackage{multicol}
\usepackage{float}
\usepackage{xcolor,colortbl}
\usepackage{amssymb}
\usepackage{fixltx2e}
\usepackage{utfsym}
\usepackage{tablefootnote}
\usepackage{tabularx}
\usepackage{ifthen} 
\usepackage{tcolorbox}
\usepackage[T1]{fontenc}
\usepackage{amsmath}
\usepackage{bbm}

\usepackage{url}

\usepackage{colortbl}
\usepackage{xcolor}
\usepackage{multirow}
\usepackage{subcaption}
\usepackage{amssymb}
\usepackage{pifont}
\usepackage{blindtext}
\usepackage{array}
\usepackage{cellspace}
\usepackage[utf8]{inputenc}
\usepackage{algorithm}
\usepackage{algpseudocode}
\usepackage{amsmath}
\usepackage{amssymb}
\usepackage{booktabs} 
\usepackage{array} 

\usepackage[table]{xcolor}
\usepackage{colortbl}
\usepackage{graphicx}
\usepackage{booktabs}
\usepackage{multirow}
\usepackage{wrapfig} 

\definecolor{mygreen}{HTML}{009900}
\definecolor{myred}{HTML}{CC0000}
\definecolor{mygray}{HTML}{666666}

\newcommand{\ours}{TARA\xspace}

\definecolor{lightgray}{rgb}{0.7,0.7,0.7}
\newcommand{\grayx}{\textcolor{lightgray}{\ding{55}}}
\newcommand{\blackcheck}{\ding{51}}

\newcommand{\figref}[2]{\ifthenelse{\equal{#2}{}}{Fig.~\ref{#1}}{Fig.~\ref{#1}\hyperref[#1]{#2}}}

\definecolor{lightergray}{HTML}{E9E9E9}
\definecolor{lighterpurple}{HTML}{E4DDF3}

\definecolor{cvprblue}{rgb}{0.21,0.49,0.74}
\usepackage[pagebackref,breaklinks,colorlinks,allcolors=cvprblue]{hyperref}

\def\confName{CVPR}
\def\confYear{2026}

\title{Taxonomy-Aware Representation Alignment for Hierarchical Visual Recognition with Large Multimodal Models}

\author{Hulingxiao He, Zhi Tan, Yuxin Peng$^*$\\
Wangxuan Institute of Computer Technology, Peking University \\
{\tt\small \texttt{hehulingxiao@stu.pku.edu.cn, tanzhi@stu.pku.edu.cn, pengyuxin@pku.edu.cn}}
}

\begin{document}
\maketitle
\insert\footins{\footnotesize$^*$Corresponding author.}
\begin{abstract} 
A high-performing, general-purpose visual understanding model should map visual inputs to a taxonomic tree of labels, identify novel categories beyond the training set for which few or no publicly available images exist. Large Multimodal Models (LMMs) have achieved remarkable progress in fine-grained visual recognition (FGVR) for known categories. However, they remain limited in hierarchical visual recognition (HVR) that aims at predicting consistent label paths from coarse to fine categories, especially for novel categories. To tackle these challenges, we propose \textbf{T}axonomy-\textbf{A}ware \textbf{R}epresentation \textbf{A}lignment (\textbf{TARA}), a simple yet effective strategy to inject taxonomic knowledge into LMMs. TARA leverages representations from biology foundation models (BFMs) that encode rich biological relationships through hierarchical contrastive learning. By aligning the intermediate representations of visual features with those of BFMs, LMMs are encouraged to extract discriminative visual cues well structured in the taxonomy tree. Additionally, we align the representations of the first answer token with the ground-truth label, flexibly bridging the gap between contextualized visual features and categories of varying granularity according to user intent. Experiments demonstrate that \ours~consistently enhances LMMs’ hierarchical consistency and leaf node accuracy, enabling reliable recognition of both known and novel categories within complex biological taxonomies. Code is available at \href{https://github.com/PKU-ICST-MIPL/TARA_CVPR2026}{https://github.com/PKU-ICST-MIPL/TARA\_CVPR2026}.
\end{abstract}    
\section{Introduction}
\label{sec:intro}

\begin{figure}[t]
    \centering
    \includegraphics[width=0.98\linewidth]{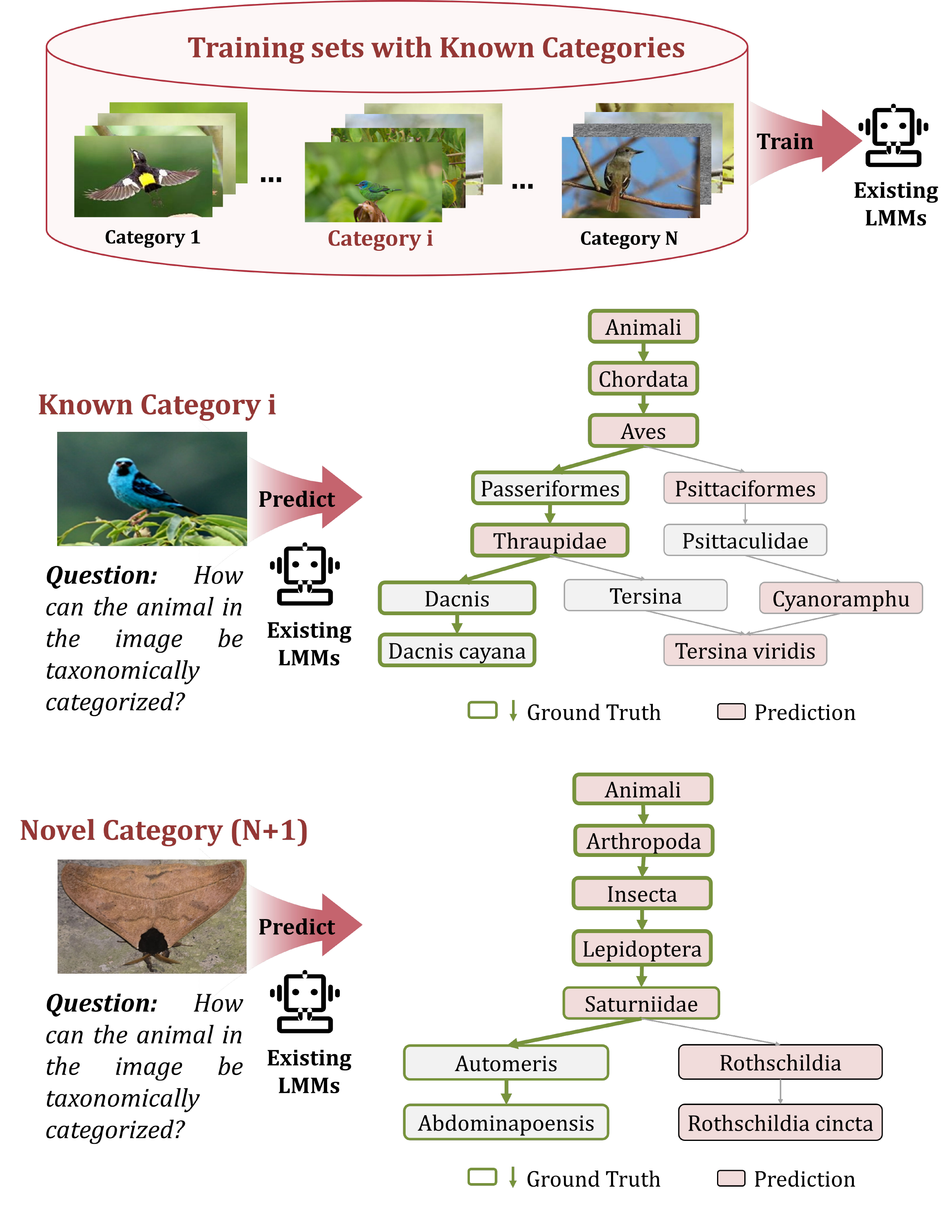}
    \caption{LMMs struggle with hierarchical visual recognition (HVR), failing to obey the hierarchical consistency on both known and novel categories.}
    \label{fig:motivation}
\end{figure}
 
Hierarchical visual recognition (HVR) \cite{chang2021fgn, hrn2022chen, biderection2024jiang, hcast2025park} aims to predict a semantic tree of labels, capturing visual categories from coarse to fine granularity. This structured output enables flexible usage: an expert user may seek a specific label such as \textit{Acadian Flycatcher}, while a general user may only require the broader category \textit{Bird}. Moreover, predicting the entire hierarchy enhances robustness and scalability, as models learn to generalize across different abstraction levels and can be easily extended to incorporate new parent or child categories.

A high-performing, general-purpose visual understanding system should not only recognize fine-grained leaf classes but also robustly map inputs to coarser, higher-level categories within a taxonomy. While existing large multimodal models (LMMs) \cite{Qwen2.5-VL,finedefics,deepperception} have demonstrated strong performance in fine-grained visual recognition (FGVR), they remain weak in hierarchical visual recognition (HVR) and often fail to maintain hierarchical consistency~\cite{tan2025vision}. As illustrated in Figure~\ref{fig:motivation}, their predictions can violate the taxonomy path—for example, breaking the sequence {Animalia} $\rightarrow$ {Chordata} $\rightarrow$ {Aves} $\rightarrow$ {Passeriformes} $\rightarrow$ {Thraupidae} $\rightarrow$ {Dacnis} $\rightarrow$ {Dacnis cayana}. Furthermore, as new categories continually emerge, LMMs must be capable of identifying novel classes that are absent from the training set and potentially lack sufficient public imagery. Because annotating data across taxonomic hierarchies requires substantial domain expertise, constructing large-scale datasets that comprehensively cover all semantic levels is infeasible~\cite{wei2021fine}. As a result, in realistic settings, LMMs often struggle to recognize novel categories.

Taxonomy is a natural and fundamental component of visual understanding. Biological taxonomies, in particular, encompass a vast range of entities that constitute our visual world~\cite{inat2021}. Recently, biological foundation models (BFMs)~\cite{bioclip,bioclip2,biocap} have emerged as large-scale repositories of biological taxonomic knowledge. For example, BioCLIP2~\cite{bioclip2} employs hierarchical contrastive learning with taxonomic supervision, leading to embedding spaces in which species representations reflect their ecological and functional relationships. In this way, visual parts~\cite{part-1,part-2,part-3}, attributes~\cite{attributes-1,attributes-2,attributes-3}, and inter-object relationships~\cite{relation-2} naturally form hierarchical groupings based on shared characteristics, providing strong priors for inferring unseen or newly discovered categories within a semantic tree.

To absorb this taxonomic knowledge within existing LMMs, we propose \textbf{T}axonomy-\textbf{A}ware \textbf{R}epresentation \textbf{A}lignment (\textbf{\ours}), a simple yet effective strategy for boosting the performance of HVR. \ours\ explicitly supervises the intermediate representations of LMMs, encouraging them to learn inter-species ecological alignment and intra-species variation from BFMs. Concretely, we first align the internal visual representations of LMMs with those of BFMs using a cosine-similarity-based alignment loss \cite{yang2025gala}. Furthermore, recognizing that a single image may correspond to multiple taxonomic levels, we additionally align the first hidden states of the output answers with the BFM-encoded category representation at a specific level. Since BFMs are trained with hierarchical contrastive objectives, they provide rich multimodal embeddings that encode strong taxonomy priors. Aligning LMM representations with those from BFMs thus enables the model to preserve fine-grained visual cues while flexibly mapping to categories of varying granularity according to user intent. Trained in an alternating manner with No-Thinking RL~\cite{cls-rl}, we demonstrate that \ours\ consistently achieves substantial performance gains across various base models, on both known and novel categories.

Our contributions are summarized as follows:
\begin{itemize}
\item We highlight a key limitation of LMMs in performing HVR, particularly for novel categories without training images, hindering general-purpose visual understanding.
\item We propose \ours, a simple but effective framework that explicitly aligns intermediate representations of LMMs with visual and text features from pretrained BFMs, thereby injecting taxonomic knowledge and enabling richer, hierarchy-aware visual recognition.
\item Through comprehensive experiments on both known and novel categories, we demonstrate consistent and significant improvements over base models, and we conduct detailed ablation studies and analyses to validate the effectiveness of each design choice.
\end{itemize}

\section{Related Work}
\label{sec:related_work}

\noindent\textbf{LMMs for FGVR.} 
Large Multimodal Models (LMMs) demonstrate strong general visual understanding but still fall short in fine-grained visual recognition (FGVR)~\cite{FOCI, zhang2024visually, finedefics, peng2025survey}, which demands distinguishing between visually similar subcategories. Several studies fine-tuned LMMs with classification-oriented data either by incorporating it during pretraining as captions or during instruction tuning as QA pairs, showing that explicit object mentions in training data are critical for accurate recognition~\cite{FOCI}. Other works suggested that the performance gap between LMMs and vision-language models (VLMs) primarily arises from data-related factors: classification-relevant cues are already encoded in the latent space but require sufficient supervision to be effectively decoded~\cite{zhang2024visually}. Integrating adequate classification data allows LMMs to approach or match specialized models and to develop stronger object-centric reasoning. From another perspective, Finedefics~\cite{finedefics} attributed this gap to misalignment between visual objects and their category names, employing attribute-based descriptions to bridge the two. Although such methods improved recognition accuracy, they lacked domain-specific objectives and explicit explanatory reasoning. To mitigate this, \cite{selfsynthx} introduced a visual rejection-sampling framework that iteratively synthesizes interpretable, feature-based explanations to enhance explainability. Moreover, Fine-R1 \cite{fine-r1} proposed a two-stage framework to learn the reasoning process with only few-shot samples per category, surpassing various strong CLIP-like models. In contrast, training-free approaches such as Sparse Attention Vectors (SAV)~\cite{SAVs} adapted LMMs for FGVR by extracting discriminative features from sparse attention heads, eliminating the need for additional training.

\noindent\textbf{Hierarchical Visual Recognition.} 
Hierarchical visual recognition~\cite{silla2011survey, kosmopoulos2015evaluation} plays a crucial role in achieving a comprehensive understanding of both the visual world~\cite{por, park2024learning, zeng2024learning, sinha2024learning, chen2022label, parkvisually} and language concepts~\cite{zhou2020hierarchy, wang2022incorporating, zhou2025novel, he2024language}. Recent research has revisited this long-standing problem, revealing that CLIP-style models~\cite{clip} often fail to maintain consistency across taxonomic levels~\cite{protect, geng2023hiclip}.
\cite{protect} evaluated CLIP under multiple levels of semantic granularity and proposed a hierarchy-consistent prompt tuning method. \cite{pal2024compositional} improved CLIP’s hierarchical representations by embedding them into a hyperbolic space, while \cite{xia2023hgclip} extended this direction with graph-based representation learning. Similarly, \cite{novack2023chils} leveraged hierarchical information to enhance zero-shot classification performance. Beyond CLIP-style models, \cite{snaebjarnarson2025taxonomy} proposed evaluating LMMs on open-set predictions \cite{zhang2024fgm} using taxonomic similarity rather than exact string matching. \cite{tan2025vision} first investigated LMMs from the perspective of HVR, and positioned the limitations of hierarchical consistency and leaf node accuracy.

\section{Preliminaries}

\subsection{Hierarchical Visual Recognition with LMMs}
In conventional visual recognition tasks, the label space is typically flat: each image ${x}\in \mathcal{X}$ is assigned a single class label $y \in \mathcal{Y}$, where $\mathcal{Y}$ denotes a predefined set of mutually exclusive categories. However, many real-world visual concepts exhibit inherent hierarchical structures, where labels are naturally organized within a taxonomy $\mathcal{T} = (\mathcal{Y}, \mathcal{E})$~\cite{parkvisually,protect,por,xia2023hgclip}, such as a tree or a directed acyclic graph (DAG). Here, $\mathcal{E} \subseteq \mathcal{Y} \times \mathcal{Y}$ represents the set of directed edges encoding parent–child relationships, where $(y_i, y_j) \in \mathcal{E}$ indicates that $y_i$ is the parent of $y_j$. In hierarchical visual recognition (HVR), the objective extends beyond predicting a single leaf-node label $y \in \mathcal{Y}_{\text{leaf}} \subseteq \mathcal{Y}$; instead, the model must also recover the complete ancestral path $(y_0, y_1, \ldots, y_L)$ in $\mathcal{T}$, where $y_0$ denotes the root node and $L$ is the hierarchy depth.

LMMs are treated as image classifiers $f_\theta$, and language prompts are leveraged to steer their outputs toward specific taxonomy levels. Concretely, we follow \cite{tan2025vision} to define a VQA-style task for each image and target taxonomy level, denoted as $(x^i, \mathcal{Y}_j)$, where $i=1,2,\ldots,N$ and $j=1,2,\ldots,L^i$. To enable evaluation in a closed-set setting, we construct four-choice VQA questions, which alleviate the challenges of open-set generation—namely, the vast output space~\cite{vlm_img_bad} and the ambiguity of prediction granularity. Generally, they follow this format \cite{tan2025vision}:
$$
\begin{array}{l}
\text{{<image> Given the plant in the image, what is its taxonomic }} \\
\text{{classification at the <hierarchy> (e.g., kingdom) level?}} \\
\text{{A.<similar class> B.<ground truth>}} \\
\text{{C.<similar class> D.<similar class>}} \\
\text{{Answer with the option letter only.}} \\
\text{(Choices are shuffled in the experiments)} 
\end{array}
$$
Although four-choice VQA tasks are arguably easier than conventional hierarchical classification, we compensate for this simplicity by designing confusing choices. Specifically, for each taxonomy level, we use SigLIP~\cite{siglip} to compute cosine similarity scores between the image and all incorrect text labels, and select the top three most similar labels as distractors. This ensures that all four options belong to the same taxonomy level and present meaningful semantic confusion.

\subsection{No-Thinking Reinforcement Fine-tuning}

Recently, rule-based reinforcement fine-tuning (RFT) has achieved remarkable progress in enhancing reasoning capabilities of LLMs (e.g., DeepSeek-R1~\cite{guo2025deepseek} and Pangu Embedded~\cite{chen2025pangu}), often surpassing traditional supervised fine-tuning (SFT) in performance~\cite{guo2025deepseek,team2025kimi,jaech2024openai}. RFT leverages verifiable rewards to guide training, encouraging models to engage in an explicit thinking process before producing answers for better solution exploration~\cite{guo2025deepseek}. This explicit reasoning is widely regarded as a key factor behind RFT’s success, and many studies on multi-modal RFT~\cite{zhou2025r1,huang2025vision} have sought to replicate the length-increasing and “aha moment” phenomena observed in DeepSeek-R1~\cite{guo2025deepseek}.

However, recent evidence suggests that RFT without explicit thinking can outperform its thinking-based counterpart on classification tasks, indicating that reasoning traces are not always necessary or beneficial—especially for smaller models~\cite{cls-rl}. Therefore, we train the base model with No-Thinking RFT as default, where instruction prompt and reward functions are as follows:

\noindent\textbf{Instruction prompt.} Unlike the Thinking-RFT prompt, which encourages step-by-step reasoning, the No-Thinking-RFT prompt explicitly prohibits the model from engaging in any thought process. It is formulated as: \texttt{\{Question\} Please directly output the answer.} 

\noindent\textbf{Reward Function.} No-Thinking-RFT removes the format reward and employs only an accuracy reward. The accuracy reward $R_{\text{accuracy}}$ assigns a value of 1 if the model’s output exactly matches the ground truth and 0 otherwise. This strict equality-based reward discourages unnecessary reasoning and enforces concise answers, significantly shorter than the reasoning-heavy outputs of Thinking-RFT. 
\section{Method}

\begin{figure*}[t]
    \centering
    \includegraphics[width=0.98\textwidth]{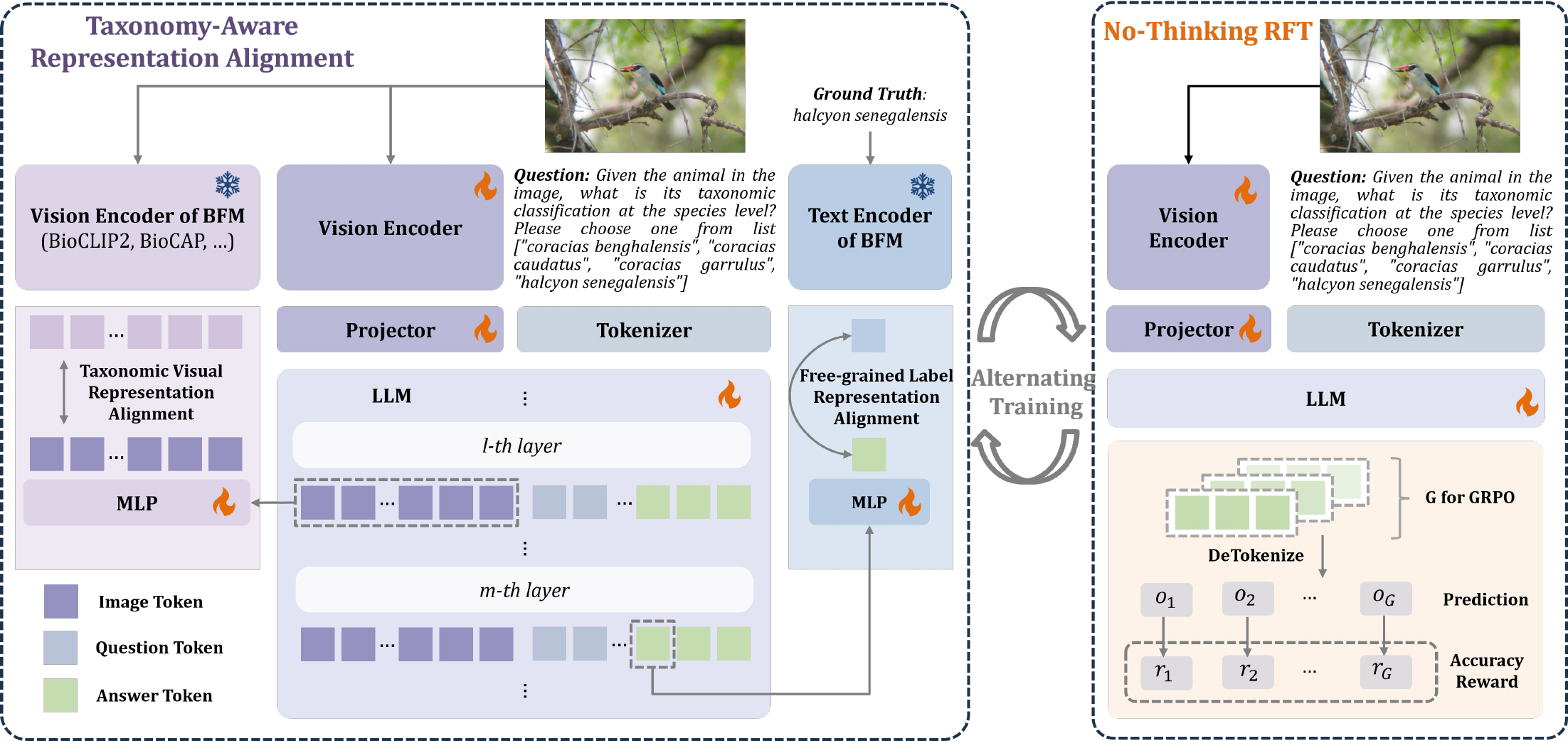}
    \caption{Illustration of the training framework. Taxonomy-Aware Representation Alignment (\ours) is conducted alternately with No-Thinking RFT to improve the hierarchical recognition performance of LMMs with taxonomic knowledge absorbed from BFMs.}
    \label{fig:pipeline}
\end{figure*}

Biology foundation models (BFMs), trained with hierarchical supervision and contrastive objectives, have demonstrated the ability to learn biologically meaningful embedding spaces~\cite{bioclip,bioclip2,biocap}. In this section, we introduce our Taxonomy-aware Representation Alignment (TARA) for LMMs. TARA aligns representations at two levels: Taxonomic Visual Representation Alignment and Free-grained Label Representation Alignment, corresponding to the specific targets of alignment. When alternately trained with No-Thinking-RFT, LMMs leverage the guidance from BFMs to enhance HVR performance larger and faster. The overall framework is illustrated in Figure~\ref{fig:pipeline}.

\subsection{Taxonomic Visual Representation Alignment}

Inspired by~\cite{viral}, we employ taxonomy-aware BFMs (e.g., BiocLIP~\cite{bioclip}, BioCLIP2~\cite{bioclip2}, and BioCAP~\cite{biocap}) as teacher models to guide the internal visual representations of LMMs. These BFMs provide rich, taxonomically informed supervision that enhances the extraction of discriminative visual cues. Concretely, we align intermediate visual features from the LMM with those produced by a pretrained BFM, thereby transferring domain-specific visual knowledge.
Let $\mathcal{E}_{\mathrm{v}}(\cdot)$ denote the pretrained BFM visual encoder. Given an input image $I$ and the question $q$ for identifying the category at a specific level (e.g., \textit{Given the animal in the image, what is its taxonomic classification at the species level? Please choose one from list [similar class, ground truth, similar class, similar class].)}, the encoder outputs target features $\mathbf{y}^{\mathrm{img}} = \mathcal{E}{\mathrm{v}}(I) \in \mathbb{R}^{N \times d}$, where $d$ is the BFM feature dimension. Let $\mathbf{e}^{\mathrm{img}}_{\ell} \in \mathbb{R}^{N \times D}$ represent the LMM’s visual representations at layer $\ell$, and let $P_{\mathrm{V}}(\cdot)$ be a learnable projection mapping $\mathbf{e}^{\mathrm{img}}_{\ell}$ into the BFM feature space. The visual alignment loss is defined as
\begin{equation}
\mathcal{L}_{\mathrm{V}} = -\frac{1}{N}\sum_{i=1}^{N}\,\mathrm{sim}\!\left(P_{\mathrm{V}}(\mathbf{e}^{\mathrm{img}}_{\ell,i}) ,\, \mathbf{y}_{i}^{\mathrm{img}}\right),
\end{equation}
where $\mathrm{sim}(\cdot)$ denotes cosine similarity and gradients do not flow into $\mathbf{y}^{\mathrm{img}}$. Minimizing $\mathcal{L}_{\mathrm{V}}$ regularizes the LMM’s internal representations, encouraging them to align with the biologically grounded visual space of the BFM.

\subsection{Free-grained Label Representation Alignment}

\begin{table*}[h]
\centering
\caption{Effects of~\ours on iNat-Plant and iNat-Animal datasets.}
\fontsize{9pt}{12pt}\selectfont  
\setlength{\tabcolsep}{6.5pt}
\begin{tabular}{ccc|ccccc|ccccc}
\toprule
\multirow{2}{*}{Base Model} & \multirow{2}{*}{RL} & \multirow{2}{*}{\ours}  & \multicolumn{5}{c|}{iNat21-Plant}&  \multicolumn{5}{c}{iNat21-Animal} \\
  \cmidrule(lr){4-8} \cmidrule(lr){9-13}  & &  & HCA & $\mathrm{Acc_{leaf}}$ & POR & S-POR & TOR  & HCA & $\mathrm{Acc_{leaf}}$ & POR & S-POR & TOR  \\
\midrule
\multirow{3}{*}{Qwen3-VL-2B} &\grayx &\grayx & 6.46 & 30.16 & 60.15 & 44.74 & 41.36 & 7.18 & 27.86 & 65.23 & 55.57 & 51.77    \\

   & \blackcheck & \grayx & 9.23 & 31.96 & 64.58 & 50.81 & 47.86 &  8.57 & 29.32 & 66.98 & 57.84 & 54.10 
   \\
        & \blackcheck & \blackcheck & \textbf{12.78} & \textbf{32.66} & \textbf{68.57} & \textbf{55.98} & \textbf{53.72} & \textbf{10.26} & \textbf{30.77} & \textbf{68.42} & \textbf{58.85} & \textbf{55.86}  
         \\
    \rowcolor{teal!10}
    & &  & \textcolor{mygreen}{+3.55} & \textcolor{mygreen}{+0.70} & \textcolor{mygreen}{+3.99} & \textcolor{mygreen}{+5.17} & \textcolor{mygreen}{+5.86} & \textcolor{mygreen}{+1.69} & \textcolor{mygreen}{+1.45} & \textcolor{mygreen}{+1.44} & \textcolor{mygreen}{+1.01} & \textcolor{mygreen}{+1.76}    \\

\midrule
 \multirow{3}{*}{Qwen2.5-VL-3B} &\grayx &\grayx & 10.89 & 39.73 & 65.77 & 48.90 & 48.24 & 16.70 & 40.26 & 73.03 & 64.05 & 60.80    \\
   & \blackcheck & \grayx & 17.91 & 44.35 & 72.15 & 57.29 & 57.08 & 21.99 & 46.25 & 76.15 & 67.66 & 64.83    \\
      & \blackcheck & \blackcheck & \textbf{19.53} & \textbf{45.66} & \textbf{73.87} & \textbf{58.92} & \textbf{59.38} & \textbf{24.02} & \textbf{49.16} & \textbf{77.26} & \textbf{68.62} & \textbf{66.11} \\
        \rowcolor{teal!10}
    & &  & \textcolor{mygreen}{+1.62} & \textcolor{mygreen}{+1.31} & \textcolor{mygreen}{+1.72} & \textcolor{mygreen}{+1.63} & \textcolor{mygreen}{+2.30} & \textcolor{mygreen}{+2.03} & \textcolor{mygreen}{+2.91} & \textcolor{mygreen}{+1.11} & \textcolor{mygreen}{+0.96} & \textcolor{mygreen}{+1.28}    \\


\bottomrule
\end{tabular}
\label{tab:benchmark_results}
\end{table*}

Unlike one-hot labels, taxonomic labels naturally encode hierarchical biological structures across multiple levels~\cite{bioclip2}. A single image may correspond to categories of varying granularity. For example, an expert might aim to identify an instance as an \textit{Acadian Flycatcher} at the species level, whereas a general user may only need the broader label \textit{Bird}. To accommodate this flexibility, we introduce free-grained label representation alignment, which aligns intermediate answer representations with the embeddings of their corresponding labels at the specified granularity level, obtained from pretrained BFMs.

Formally, the BFM encoder $\mathcal{E}_{\mathrm{T}}(\cdot)$ generates a target feature $\mathbf{y}^{\mathrm{label}} = \mathcal{E}_{\mathrm{T}}(C) \in \mathbb{R}^{d}$, where $C$ denotes the ground-truth label at the desired granularity. Let $\mathbf{e}^{\mathrm{answer}}_{m} \in \mathbb{R}^{N'\times D}$ represent the answer embeddings from the LMM at layer $m$. We then employ a projector $P_{\mathrm{T}}(\cdot)$ to map the first-token embeddings of $\mathbf{e}^{\mathrm{answer}}_{m}$ into the BFM textual feature space. The alignment loss is defined as:
\begin{equation}
\mathcal{L}_{\mathrm{C}} = - \mathrm{sim}\left(P_{\mathrm{T}}(\mathbf{e}^{\mathrm{answer}}_{m,0}), \mathbf{y}^{\mathrm{label}}\right),
\end{equation}
where $\mathbf{e}^{\mathrm{answer}}_{m,0}$ denotes the embedding of the first token in the generated answer. Minimizing $\mathcal{L}_{\mathrm{C}}$ encourages the answer representations to align with the ground-truth labels at the appropriate semantic level, thereby enabling the model to learn structured, hierarchy-aware representations beneficial for downstream HVR tasks. The overall alignment objective of \ours~is given by:
\begin{equation}
\mathcal{L}_\mathrm{alignment} = (\mathcal{L}_{\mathrm{V}} +  \mathcal{L}_{\mathrm{C}}) / 2.
\end{equation}

\begin{algorithm}[t]
\begingroup
\setlength{\lineskip}{0pt}
\setlength{\lineskiplimit}{0pt}

\caption{No-Thinking RFT + TARA}
\label{alg:tara-compact}

\begin{algorithmic}[1]
\Require input image $I$, question $q$, label $C$; LMM $\pi_\theta$; BFM encoders $\mathcal{E}_{\rm V},\mathcal{E}_{\rm T}$; projectors $P_{\rm V},P_{\rm T}$
\Ensure Updated LMM $\pi_\theta$

\For{each step}
    \State Generate $G$ answers; extract $e^{\rm img}_\ell, e^{\rm answer}_{m,0}$.
    \State Compute $\mathcal{L}_{\rm RFT}$.
    \State Encode GT: $\mathbf{y}^{\rm img}=\mathcal{E}_{\rm V}(I)$, $\mathbf{y}^{\rm label}=\mathcal{E}_{\rm T}(C)$.
    \State Project: $\hat{\mathbf{y}}^{\rm img}=P_{\rm V}(e^{\rm img}_\ell)$, $\hat{\mathbf{y}}^{\rm label}=P_{\rm T}(e^{\rm answer}_{m,0})$.
    \State Compute $\mathcal{L}_{\rm alignment}$.
    \State Update $\pi_\theta,P_{\rm V},P_{\rm T}$ with $\mathcal{L}_{\rm alignment}$.
    \State Update $\pi_\theta$ with $\mathcal{L}_{\rm RFT}$ (with $P_{\rm V},P_{\rm T}$ frozen).
\EndFor

\end{algorithmic}
\endgroup
\end{algorithm}

\subsection{Training and Inference}
Algorithm \ref{alg:tara-compact} shows the training process. In addition to applying No-Thinking RFT to optimize LMMs, we integrate \ours~to jointly update the LMMs and two lightweight MLP-style projectors, $P_{\mathrm{V}}(\cdot)$ and $P_{\mathrm{T}}(\cdot)$, which map intermediate representations of the LMM. This alternating optimization scheme enables the model to effectively absorb taxonomic knowledge, leading to improved performance and generalization in HVR. During inference, both the BFMs and projectors are discarded, and LMMs are directly prompted to perform HVR.
\section{Experiments}
\label{sec:exp}

\subsection{Experimental Setup}

\noindent\textbf{Datasets.} 
We employ the iNaturalist-2021 (iNat21) dataset~\cite{inat2021}, a large-scale collection featuring species-level annotations across diverse biological taxa. We separate it into two taxonomies, Plant and Animal, comprising 4,271 and 5,388 leaf nodes, respectively, across six hierarchical levels.
To assess the model’s ability to recognize unseen species, we further incorporate the TerraIncognita dataset~\cite{terraincognita}. This dataset includes a mixture of expertly annotated images of insect species that are likely familiar to frontier AI models, as well as images of rare or poorly documented species with few or no publicly available samples. These novel-category images were collected during field expeditions in biodiversity hotspots across Central and South America. For many of these samples, only higher-level taxonomic labels are reliable, and numerous taxa are believed to be entirely new to science.

\noindent\textbf{Implementation Details.}
We adopt the Qwen family of models as base models due to their strong zero-shot performance. Specifically, we employ Qwen3-VL-2B-Instruct~\cite{qwen3technicalreport} and Qwen2.5-VL-3B-Instruct~\cite{Qwen2.5-VL} as our base models, and fine-tune all parameters following the training configurations of~\cite{zhou2025r1, chen2025r1v}.
Our No-Thinking RFT dataset is constructed from the iNat21-Animal and iNat21-Plant training splits, containing 1-shot VQA samples across 9,659 leaf categories. For hierarchical recognition, we formulate questions corresponding to the \emph{order}, \emph{family}, \emph{genus}, and \emph{species} levels, sampled at a ratio of 1:2:4:8. Evaluation is conducted using 1-shot VQA samples derived from the corresponding iNat-Animal and iNat-Plant validation sets. The vision and text projectors, $P_\mathrm{V}(\cdot)$ and $P_\mathrm{T}(\cdot)$, are implemented as lightweight three-layer MLPs with SiLU activations, while $\mathcal{E}_\mathrm{V}(\cdot)$ and $\mathcal{E}_\mathrm{T}(\cdot)$ are the vision and text encoders of BioCLIP2~\cite{bioclip2}, respectively. All experiments are performed on 8×A6000 GPUs with a batch size of 1 per GPU and a two-step gradient accumulation~\cite{chen2025r1v, shen2025vlm}. Each model is trained for one epoch. We adopt the few-shot classification hyper-parameters from~\cite{cls-rl}. All input images are resized to 328×328, and no data augmentation is applied.

\subsection{Evaluation Metrics}

To comprehensively evaluate model performance, we focus on the hierarchical consistency of predictions~\cite{protect, park2024learning}, complemented by the leaf-level classification accuracy~\cite{vlm_img_bad, liu2024revisiting, finedefics}, which serves as the upper bound of hierarchical consistency. The evaluation metrics are detailed below.

\noindent\textbf{Hierarchical Consistent Accuracy (HCA)} \cite{protect,park2024learning}. This metric is defined as 
\begin{align}
    \mathrm{HCA} = \frac{1}{N} \sum_{i=1}^{N} \prod_{j=1}^{L^i} \mathbbm{1}\left[f_\theta\left(x^i; \mathcal{Y}_{j}\right) = y^i_j\right], \label{eq:HCA}
\end{align}
Here, $N$ is the number of test samples, $L^i$ is the depth of the hierarchy for the $i$-th input $x^i$, and $\mathcal{Y}{j}$ denotes the label set at level~$j$. HCA computes the proportion of samples whose predicted paths exactly match the ground truth from root to leaf. It is therefore a stricter criterion than flat accuracy and serves as our primary evaluation metric for hierarchical classification.

\begin{table}[t]
\centering
\caption{Effects of \ours~on TerraIncognita dataset~\cite{terraincognita}.}
\fontsize{9pt}{12pt}\selectfont
\setlength{\tabcolsep}{10pt}
\begin{tabular}{ccc|cc}
\toprule
\multirow{1}{*}{Species}  & \multirow{1}{*}{RL} & \multirow{1}{*}{\ours}  & \multicolumn{1}{c}{Order F1}&  \multicolumn{1}{c}{Family F1}  \\
\midrule
\multirow{3}{*}{Known} &\grayx &\grayx & 17.16 & 10.83  \\
  & \blackcheck & \grayx & 23.30 & 11.47
   \\
      & \blackcheck & \blackcheck & 41.56 & 25.47 \\
        \rowcolor{teal!10}
  &    &   & \textcolor{mygreen}{+18.26} & \textcolor{mygreen}{+14.00} \\


\midrule
\multirow{4}{*}{Novel}  &\grayx &\grayx & 17.16 & 10.83 \\
  & \blackcheck & \grayx & 23.30 & 11.47 
   \\
       & \blackcheck & \blackcheck & 33.45 & 12.67\\
        \rowcolor{teal!10}
  &    &   & \textcolor{mygreen}{+10.15} & \textcolor{mygreen}{+1.20} \\



\bottomrule
\end{tabular}
\label{tab:benchmark_results}
\end{table}

\noindent\textbf{Leaf-Level Accuracy ($\mathrm{Acc_{leaf}}$)}~\cite{vlm_img_bad, liu2024revisiting, finedefics}.
It reflects the model’s discriminative ability at the most fine-grained level:
\begin{align}
    \mathrm{Acc_{leaf}} = \frac{1}{N} \sum_{i=1}^{N} \mathbbm{1}\left[f_\theta\left(x^i; \mathcal{Y}_{L}\right) = y^i_L\right]. \label{eq:Acc}
\end{align}
Since correctly predicting a leaf node implies correctness at that level, $\mathrm{Acc_{leaf}}$ naturally upper-bounds HCA. However, a sample contributes to HCA only if all nodes along its path $(y_0, y_1, \dots, y_L)$ are predicted correctly, making $\mathrm{HCA}$ a more stringent measure of hierarchical consistency.

\noindent\textbf{Point-Overlap Ratio (POR)}~\cite{por}.
It measures hierarchical performance beyond strict correctness, defined as:
\begin{equation}
    \mathrm{POR} = \frac{1}{N} \sum_{i=1}^{N} \frac{\sum_{j=1}^{L_i} \mathbbm{1}\left[f_\theta\left(x_i; \mathcal{Y}_{j}\right) = y^i_j\right]}{L_i}.
\end{equation}
Unlike HCA, which requires an exact match along the entire path, POR allows partial correctness by averaging the proportion of correctly predicted nodes. This provides a fine-grained assessment of how well model outputs align with the target hierarchy.

\noindent\textbf{Strict Point-Overlap Ratio (S-POR).}
S-POR refines POR by rewarding only contiguous segments of correct predictions. For the $i$-th sample, we locate the longest run of consecutive correctly predicted nodes and normalize by the hierarchy depth $L_i$:
\begin{align}
\mathrm{S\text{-}POR}
&= \frac{1}{N}\sum_{i=1}^{N}\frac{1}{L_i}
  \max_{1\le a\le b\le L_i}
  \Bigl[(b-a+1)
  \notag\\[-3pt]
&\qquad\times
        \prod_{j=a}^{b}
        \mathbbm{1}\bigl[f_\theta(x_i;\mathcal{Y}_{j}) = y^i_j\bigr]\Bigr].
\end{align}
This stricter definition penalizes isolated correct predictions and emphasizes full-path consistency within the hierarchy.

\noindent\textbf{Top Overlap Ratio (TOR).}
Following \cite{protect}, TOR evaluates local hierarchical consistency by considering adjacent layer pairs as independent evaluation units:
\begin{equation}
\begin{aligned}
\mathrm{TOR}
&= \frac{1}{N}\sum_{i=1}^{N}
    \frac{1}{L_i-1}\sum_{j=1}^{L_i-1}
    \mathbbm{1}\bigl[f_\theta(x_i;\mathcal{Y}_{j}) = y^i_j\bigr]
\\[-3pt]
&\quad\times
    \mathbbm{1}\bigl[f_\theta(x_i;\mathcal{Y}_{j+1}) = y^i_{j+1}\bigr].
\end{aligned}
\label{eq:tor}
\end{equation}
A TOR value of~1 indicates perfect pairwise consistency between consecutive layers, while lower scores reveal local violations of the hierarchical structure.

\noindent\textbf{F1 Score.} For evaluations on TerraIncognita, we follow the setup of~\cite{terraincognita} and use a fixed system prompt instructing the model to classify an insect specimen across the taxonomic hierarchy, returning “Unknown’’ when it is uncertain. We report F1 score at the Order and Family level since full taxonomic labels are missing for novel species. It captures the harmonic mean of precision and recall and thus provides a balanced assessment of the model’s classification performance.

\begin{table}[t]
\centering
\caption{Ablation on target alignment layers of \ours.}
\fontsize{9pt}{12pt}\selectfont
\setlength{\tabcolsep}{6pt}
\begin{tabular}{cc|ccccc}
\toprule
\multicolumn{2}{c|}{Target Layer}  & \multicolumn{5}{c}{iNat21-Plant} \\
 \cmidrule(lr){1-2} \cmidrule(lr){3-7}  {$\mathcal{L}_\mathrm{V}$} & {$\mathcal{L}_\mathrm{T}$} & HCA & $\mathrm{Acc_{leaf}}$ &  POR & S-POR & TOR   \\
\midrule
/ & / & 6.46 & 30.16 & 60.15 & 44.74 & 41.36  \\
14 & / & 10.72 & 33.27 & 65.35 & 51.40 & 49.11  \\
14 & 14 & 10.40 & \textbf{33.43} & 65.27 & 51.14 & 49.26  \\
14 & 28 & \textbf{12.78} & 32.66 & \textbf{68.57} & \textbf{55.98} & \textbf{53.72}  \\
28 & 14 & 11.87 & 33.20 & 67.17 & 54.62 & 51.88 \\
28 & 28 & 10.65 & 32.73 & 66.52 & 53.41 & 50.48 \\

\bottomrule
\end{tabular}
\label{tab:ablation_core_components}
\end{table}

\begin{figure*}[t]
    \centering
    \includegraphics[width=0.98\linewidth]{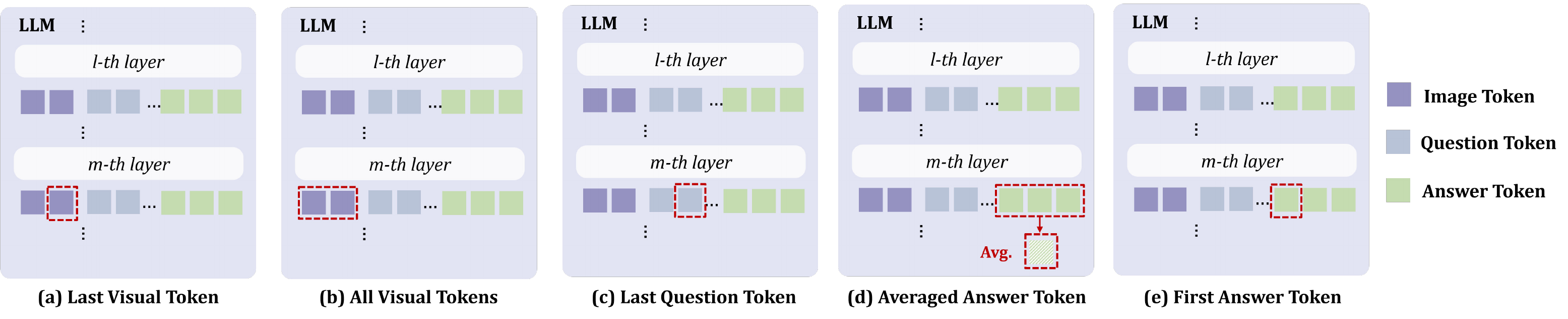}
    \caption{Different designs of target alignment features. (a) and (b) are for $\mathcal{L}_\mathrm{V}$, and (c)-(e) are for $\mathcal{L}_\mathrm{C}$.}
    \label{fig:text_alignment_design}
\end{figure*}

\subsection{Main Results}

\noindent\textbf{Known Categories.} The results on iNat-Animal and iNat-Plant are summarized in Table~\ref{tab:benchmark_results}. With only 1-shot supervision, both base models (Qwen3-VL-2B and Qwen2.5-VL-3B) trained with \ours~consistently outperform their baselines, achieving improvements in both hierarchical consistency and leaf-level accuracy. This is obtained through a simple yet effective strategy that aligns intermediate LMM features with BFM targets, enabling the model to absorb taxonomic structure. On the known split of the TerraIncognita dataset~\cite{terraincognita}, \ours~also delivers substantial gains on Order F1 and Family F1. These consistent improvements demonstrate that \ours~effectively guides LMMs to recognize known categories more reliably.

\noindent\textbf{Novel Categories.} To assess whether these gains stem merely from memorizing known training categories, we further evaluate on the novel split of the TerraIncognita dataset~\cite{terraincognita}, which contains newly captured images of rare or previously unseen species beyond the scope of training data. Even in this challenging setting, \ours~provides consistent improvements, indicating that the learned representations generalize beyond observed classes within the tree of life. Overall, these results highlight a broader insight: regularizing intermediate representations is a broadly applicable approach for strengthening LMMs with taxonomic knowledge, even under extremely limited data.

\begin{table}[t]
\centering
\caption{Ablation on target alignment features of \ours.}
\fontsize{9pt}{12pt}\selectfont
\setlength{\tabcolsep}{7pt}
\begin{tabular}{cc|ccccc}
\toprule
\multicolumn{2}{c|}{Features}  & \multicolumn{5}{c}{iNat21-Plant} \\
 \cmidrule(lr){1-2} \cmidrule(lr){3-7}  {$\mathcal{L}_\mathrm{V}$} & {$\mathcal{L}_\mathrm{T}$} & HCA & $\mathrm{Acc_{leaf}}$ &  POR & S-POR & TOR   \\
\midrule
(a) & (e) &  10.30 & 32.22 & 65.50 & 52.36 & 49.59 \\
(b) & (c) &  12.01 & \textbf{33.62} & 67.39 & 55.12 & 52.25 \\
(b) & (d) &  11.59 & 33.41 & 67.18 & 53.50 & 51.53 \\
(b) & (e) & \textbf{12.78} & 32.66 & \textbf{68.57} & \textbf{55.98} & \textbf{53.72}  \\  
\bottomrule
\end{tabular}
\label{tab:ablation_design}
\end{table}

\subsection{Component-wise Analysis}

In this ablation study, we systematically analyze the key design choices of our framework, examining the contribution of each core component as well as the impact of different target alignment features and the selected layers used in the alignment losses.

\noindent\textbf{Effects of core components.} We first analyze the contribution of each alignment loss introduced in \ours, namely $\mathcal{L}_\mathrm{V}$ and $\mathcal{L}_\mathrm{C}$. As shown in Table~\ref{tab:ablation_core_components}, incorporating $\mathcal{L}_\mathrm{V}$ enhances both hierarchical consistency and leaf-level accuracy, indicating that the alignment signal from the BFM vision encoder carries taxonomy-aware visual knowledge. Introducing $\mathcal{L}_\mathrm{C}$ further improves the intermediate LMM representations, helping the model better map an input image to labels across different levels of granularity. This leads to consistent improvements in hierarchical consistency (e.g., a 2.06\% gain in HCA.)

\noindent\textbf{Target alignment layers.} We then analyze alignment at individual target layers to determine the most effective positions of $\mathcal{L}_\mathrm{V}$, $\mathcal{L}_\mathrm{C}$. As shown in the target-layers ablation results in Table~\ref{tab:ablation_core_components}, we report performance at different layers throughout the network. We observe that performance varies depending on the alignment layer, with the 14th, 28th layer of the 28-layer model consistently yielding stronger results. This trend is consistent with the intuition that the visual information is progressively transferred to taxonomic label and finally to the label at a certain level according to the given question.

\noindent\textbf{Target alignment features.}
For the visual alignment loss $\mathcal{L}_\mathrm{V}$, we compare two design choices for constructing target alignment features: using only the embedding of the last visual token (Figure~\ref{fig:text_alignment_design} (a)) versus using the embeddings of all visual tokens (Figure~\ref{fig:text_alignment_design} (b)).
For the text alignment loss $\mathcal{L}_\mathrm{C}$, we evaluate three alternatives: using the embedding of the last question token (Figure~\ref{fig:text_alignment_design} (c)), the averaged embeddings of all answer tokens (Figure~\ref{fig:text_alignment_design} (d)), and our proposed approach, which uses the embedding of the first answer token (Figure~\ref{fig:text_alignment_design} (e)).
As summarized in Table~\ref{tab:ablation_design}, the “All Visual Tokens’’ variant yields the best performance for $\mathcal{L}_\mathrm{V}$, while the “First Answer Token’’ strategy is the most effective choice for $\mathcal{L}_\mathrm{C}$.

\begin{table}[t]
\centering
\caption{Left: visual probing results on the last and averaged image hidden states. Right: evaluation results on ImageWikiQA~\cite{zhang2024visually}.}
\begin{minipage}{0.58\linewidth}
\centering
\fontsize{9pt}{12pt}\selectfont
\setlength{\tabcolsep}{7pt}
\begin{tabular}{cc|cc}
\toprule
RL & \ours & Last & Avg. \\
\midrule
\grayx &\grayx & 13.30 & 85.00  \\
\blackcheck & \grayx & 14.40 & 84.40 \\
\blackcheck & \blackcheck & 18.30 & 85.90 \\
\rowcolor{teal!10}
 & & \textcolor{mygreen}{+3.90} & \textcolor{mygreen}{+1.50} \\
\bottomrule
\end{tabular}
\end{minipage}
\hfill
\begin{minipage}{0.4\linewidth}
\centering
\fontsize{9pt}{12pt}\selectfont
\setlength{\tabcolsep}{6pt}
\begin{tabular}{cc|c}
\toprule
RL & \ours & Acc. \\
\midrule
\grayx &\grayx & 46.60  \\
\blackcheck & \grayx & 48.70  \\
\blackcheck & \blackcheck & 51.40 \\
\rowcolor{teal!10}
 &  & \textcolor{mygreen}{+2.70} \\
\bottomrule
\end{tabular}
\end{minipage}
\label{tab:visual_probing}
\end{table}





\subsection{Probing Analysis}

To examine whether \ours~enhances visual representations by enabling the extraction of more discriminative cues for HVR, we conduct linear probing experiments on the image token embeddings from the residual stream at the last layer of the LLM. Two pooling strategies are explored: mean pooling across all image tokens and selecting the final image token. A linear classifier is trained on the iNat21-Plant training set with a batch size of 512, a learning rate of 1e-4, and the Adam optimizer for 500 epochs. To ensure balance, we randomly sample 10 images per class from 100 categories (1,000 images total) for training, and use another 1,000 images for testing. Table~\ref{tab:visual_probing} (left) reports species-level classification accuracy. We observe that \ours~outperforms No-Thinking RFT variant, demonstrating its superior ability to extract fine-grained visual cues.

\subsection{Classification-based VQA Benchmark}
Classification serves as a fundamental building block for developing more advanced visual capabilities. For instance, correctly recognizing an object is often a prerequisite for answering complex questions about it. To demonstrate that \ours~extends beyond HVR and benefits more challenging tasks, we further evaluate its performance on ImageWikiQA~\cite{zhang2024visually}, a dataset containing complex, real-world questions grounded in ImageNet objects. As shown in Table~\ref{tab:visual_probing} (right), \ours~improves accuracy from 48.70\% to 51.40\%, highlighting that strengthening HVR indeed enhances the advanced reasoning capabilities of LMMs.

\subsection{Qualitative Results}
\begin{figure}[t]
    \centering
    \includegraphics[width=0.98\linewidth]{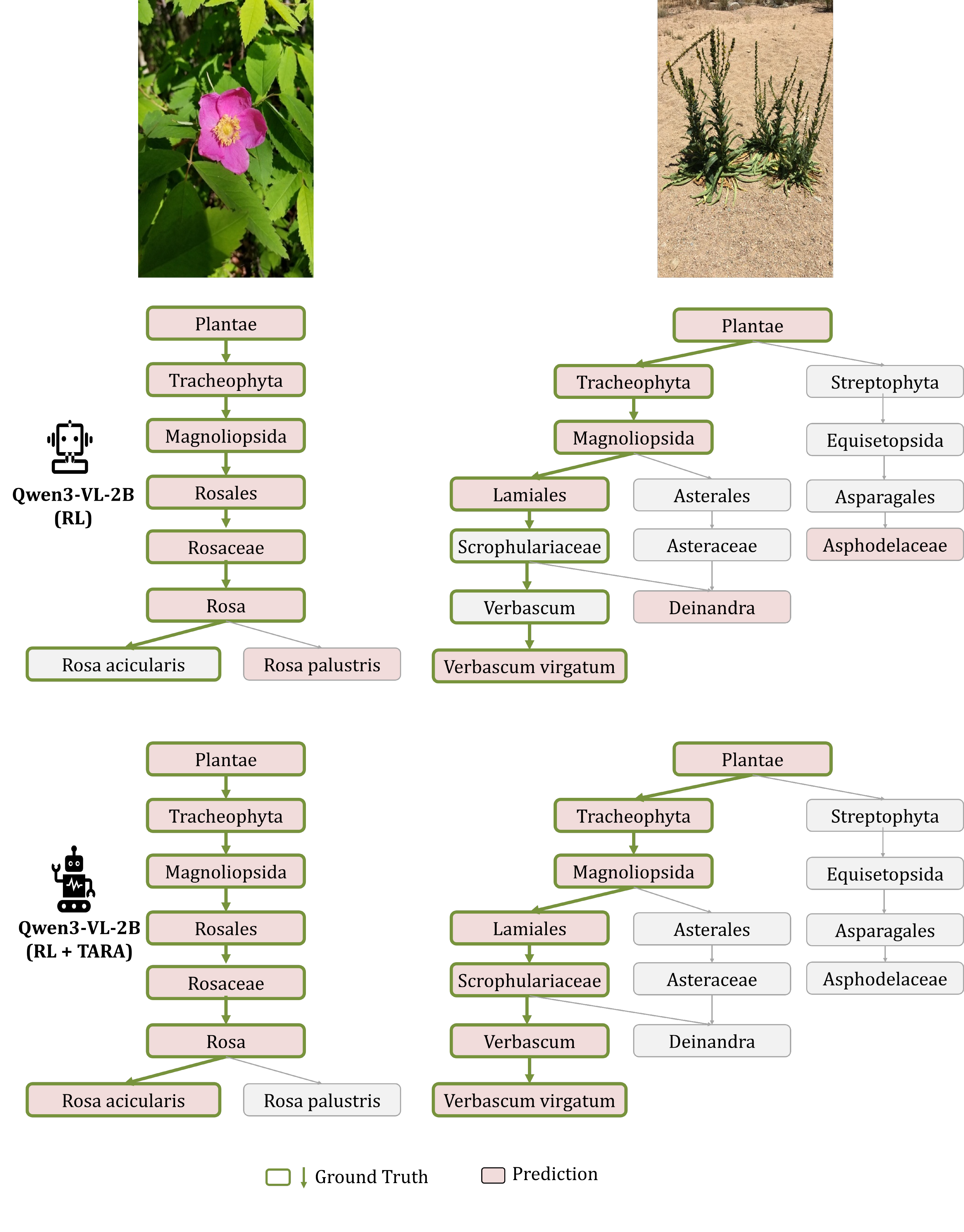}
    \caption{Qualitative comparison of No-Thinking RFT with and without \ours. The two columns show that \ours~can achieve better leaf node accuracy and hierarchical consistency.}
    \label{fig:case}
\end{figure}

We further validate the effectiveness of \ours~through qualitative analyses of model outputs. As shown in Figure~\ref{fig:case}, \ours~not only improves fine-grained prediction accuracy but also rectifies errors along the entire hierarchical label path, yielding substantially better hierarchical consistency than merely applying No-Thinking RFT to Qwen3-VL-2B.

\subsection{Training Efficiency}
\begin{figure}[t]
    \centering
    \includegraphics[width=0.98\linewidth]{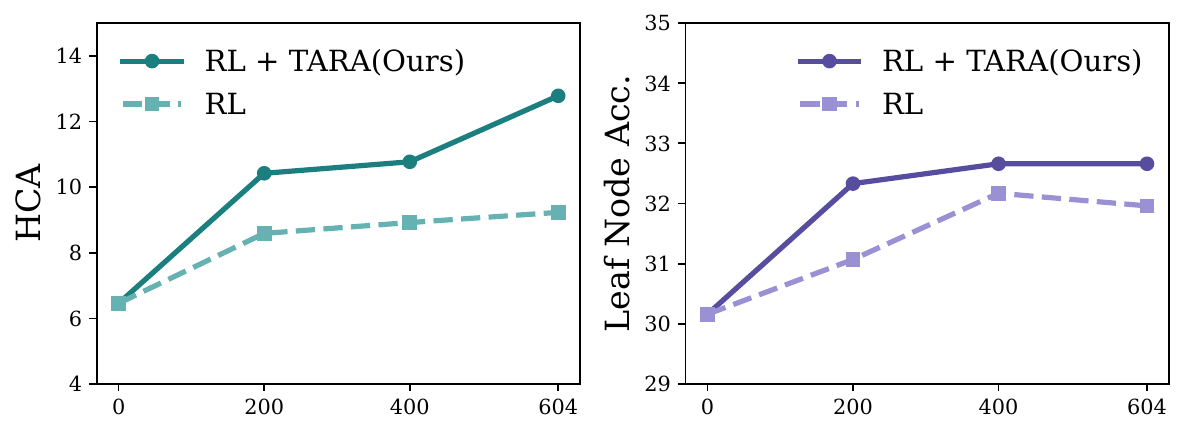}
    \caption{Training efficiency. Models trained with \ours~achieve faster convergence.}
    \label{fig:training_efficiency}
\end{figure}

To further showcase the additional benefits of \ours, we evaluate iNat21-Plant at every 200 training step from the total of 604 training steps of the No-Thinking RFT stage in Figure~\ref{fig:training_efficiency}. Qwen3-VL-2B trained with \ours~show that convergence is faster and quickly surpasses the baseline performance in early steps, demonstrating the effectiveness of representation guidance for taxonomy knowledge injection. Since \ours~adds minimal overhead to the overall training process, these earlier performance gains are likely to translate into enhanced scalability.

\section{Conclusion}
\label{sec:con}

In this work, we introduce \ours, a simple yet effective strategy that aligns the internal representations of LMMs with those of pre-trained BFMs. By doing so, our approach enables the extraction of fine-grained visual semantics, facilitating accurate mapping to taxonomic labels and allowing the transfer of visual features across categories at any level of the tree of life. Extensive experiments demonstrate that \ours improves recognition of known categories and generalizes effectively to novel categories in the HVR task.

\noindent\textbf{Limitations.}
More real-world classification tasks beyond the biological domain involve hierarchical label spaces (e.g., taxonomies or knowledge graphs). Incorporating such structure could further advance LMMs toward general-purpose visual understanding.

\section*{Acknowledgements}
This work was supported by the grants from the National Natural Science Foundation of China (62525201, 62132001, 62432001) and Beijing Natural Science Foundation (L247006). This work was partially supported by PKU Kunpeng\&Ascend Center of Excellence.

\vspace{-2mm}
{
    \small
    \bibliographystyle{ieeenat_fullname}
    \bibliography{main}
}

\end{document}